\documentclass[11pt]{article}

\usepackage[final]{acl}

\usepackage{times}
\usepackage{latexsym}

\usepackage[T1]{fontenc}

\usepackage[utf8]{inputenc}

\usepackage{microtype}

\usepackage{inconsolata}

\usepackage{graphicx}
\usepackage{booktabs}
\usepackage{multirow}
\title{External Experience Serving in Production LLM Systems: A Deployment-Oriented Study of Quality-Cost Trade-offs}

 \author{Lin Sun \and Heming Zhang \\
         Qiyuan Tech \\
\small{
    \textbf{Correspondence:} \href{mailto:sunlin1@360.cn}{sunlin1@360.cn}}
}

\begin{document}
\maketitle
\begin{abstract}
Production LLM systems accumulate reusable operational experience, but serving it at inference time introduces quality–cost trade-offs. We compare no experience, random injection, static global injection, and request-conditioned retrieval on a production-derived moderation benchmark, with tool routing and GPQA-diamond as contrasting workloads. We measure task quality and full request-path latency, including retrieval and prompt assembly. Retrieval achieves 82.1\% fine-grained moderation accuracy, compared with 46.7\% without experience and 69.4\% with a compact global prompt. Relative to the compact prompt, retrieval increases latency from 0.30 s to 0.58 s and benign false-positive rate from 5.2\% to 7.1\%. The moderation accuracy gains extend to additional model families. Retrieval increases latency in short-output workloads; on GPQA-diamond in the Instruct regime, however, shorter completions coincide with lower latency (19.35 s to 16.92 s) and higher observed accuracy. In moderation, enabling prefix caching reverses the latency ordering of compact global injection and retrieval. Together with retrieval-budget sweeps, these results show that quality gains alone do not determine the preferred serving strategy: cache reuse, generation length, and retrieval budget must be assessed against task-specific quality, false-positive, and latency constraints.
\end{abstract}

\section{Introduction}

Production LLM systems accumulate reusable operational experience, including prior edge cases, policy refinements, resolution traces, and expert-authored rules. The practical question is not merely whether this experience can help at inference time, but how different serving choices trade off quality against online cost under realistic deployment constraints. External experience can improve decision quality without retraining model weights~\citep{ram-etal-2023-context}, but processing additional context can increase online serving cost~\citep{kwon2023efficientmemorymanagementlarge,dao2022flashattentionfastmemoryefficientexact}. We therefore study \textit{external experience serving} as a deployment-oriented quality-cost trade-off problem.

We focus on serving-time use of an existing experience repository rather than on repository construction, maintenance, or parameter-level internalization. Accuracy, prompt length, and full request-path latency can rank serving strategies differently. We therefore ask how static global injection compares with selective retrieval, and how cache reuse, generation workload, and retrieval budget change which configurations merit deployment consideration.

Our primary deployment case is moderation\citep{openai_moderation_guide}, where errors are costly, policies evolve over time, and reusable operational experience naturally accumulates. Tool-use (tool routing) and GPQA provide short-output and decode-heavy contrasts, respectively. Across tasks, we evaluate paired Instruct and Thinking regimes so that the same task can move to a different cost-quality operating point when decode burden changes substantially.

We make three contributions through a deployment-oriented study of external experience serving. First, we compare serving interfaces in a production moderation workload and two contrast tasks, tool-use and GPQA-diamond, measuring task quality and full request-path latency. Second, we locate Moderation gains in fine-grained hierarchical decisions, quantify benign false positives, and replicate overall accuracy improvements across additional model families. Third, we show how serving conditions change the relative value of these interfaces: prefix caching reverses the latency ordering of Global compact and Retrieval, while shorter GPQA completions coincide with lower Full E2E despite larger prompts. Budget sweeps further distinguish configurations that improve both observed accuracy and latency from those that exchange small quality gains for substantial latency. Together, these comparisons explain why experience-serving strategies should be selected under explicit quality and service constraints.

\section{Related Work}

Retrieval-augmented generation and external-context methods incorporate knowledge beyond model parameters~\citep{lewis2021retrievalaugmentedgenerationknowledgeintensivenlp,ram-etal-2023-context}. Related studies examine context selection, budgets, and long-input sensitivity~\citep{shi2023replugretrievalaugmentedblackboxlanguage,liu-etal-2024-lost,sun2026bearbudgetedevidenceallocation}.

RAGCache caches retrieved knowledge states and overlaps retrieval with inference~\citep{jin2025ragcache}. CacheBlend reuses non-prefix KV caches with selective recomputation to preserve generation quality~\citep{yao2025cacheblend}. Cache-augmented generation (CAG) preloads and caches manageable knowledge collections as an alternative to online retrieval~\citep{chan2025dontdorag}.

We compare static and request-conditioned experience serving using fine-grained Moderation quality, benign false positives, and full request-path latency. Tool-use and GPQA provide workload contrasts; cache and budget measurements assess configuration sensitivity under the evaluated prefix-cache implementation. Our static baselines use fixed subsets or compressed summaries, differing from CAG's full-collection preloading. These comparisons inform experience-serving policy choices under application-specific quality, false-positive, and latency constraints.

\section{Experience Serving in Production}

\subsection{Experience sources}

We study the inference-time use of existing repositories of operational experience accumulated during deployment. Such experience may take the form of expert-authored policy bullets, prior failure cases, escalation resolutions, stable decision rules, or structured edge-case traces. Such repositories may also be refined or expanded by reflective and agentic context-construction pipelines explored in recent work~\citep{zhang2026agenticcontextengineeringevolving,agrawal2026gepareflectivepromptevolution}. Throughout the paper, we use \textit{experience item} to denote one such reusable unit of experience.

\subsection{Online experience serving interface}

We compare two interfaces for serving an existing experience repository. Static global experience injection inserts a fixed experience block into every request. Global full uses a fixed set of complete experience items, while Global compact uses a compressed shared summary. Both can benefit from prefix-cache reuse, although the fixed block may contain guidance that is irrelevant to a particular request.

Request-conditioned retrieval, referred to as Retrieval, selects experience for the current request and injects the selected items into the prompt. The online pipeline forms a request-specific query, combines dense and lexical candidates, fuses and reranks results, and assembles the Top-$K$ experience block before generation. It can provide more targeted guidance, but introduces retrieval and prompt-assembly overhead and can reduce prefix-cache reuse. The deployment question is whether request-conditioned retrieval offers a preferable quality--cost operating point to static global injection under the target task and serving constraints. We evaluate this comparison using task quality and full request-path latency; retrieval implementation and prompt-assembly details are provided in Appendices~\ref{sec:appendix-retrieval-details} and~\ref{sec:appendix-prompt-schema}.

\subsection{Task-dependent serving regimes}

Even after the serving interface is fixed, the value of external experience remains task-dependent. Retrieval and prompt assembly introduce additional online work, whose latency impact depends on the workload and serving configuration. When completions are long, changes in generation length can also affect the total cost of serving experience.

We therefore analyze quality--cost trade-offs across tasks and inference regimes. Moderation and Tool-use provide relatively short-output workloads, while GPQA provides a decode-heavy comparison. Within each task, Instruct and Thinking regimes vary the generation workload, allowing us to examine how experience serving changes task quality and full request-path latency under each configuration.

\section{Experimental Setup}

\subsection{Tasks and deployment setting}

Our primary task is a real production moderation workload for LLM safety guarding, where failures are costly, policies evolve over time, and reusable operational experience accumulates naturally. The live system processes approximately 12B tokens per day. The task is formulated as a hierarchical multi-class safety categorization problem with 16 top-level categories, 64 second-level categories, and 94 third-level categories. The taxonomy has a maximum depth of three. Each example is annotated with a terminal leaf label at L1, L2, or L3; benign is an L1 leaf. The gold classification path extends from L1 to that terminal label. The Moderation evaluation set contains 1,500 labeled examples: 934 risk examples and 566 benign examples. Labels are initialized by an LLM, cross-annotated by three human annotators, and verified by a quality inspector. Annotation-consistency checks exceed 95\%; further protocol details appear in Appendix~\ref{sec:appendix-moderation-protocol}.

Tool-use (800 examples) evaluates tool routing on a production-derived workload. Given a user request and descriptions of the available tools, the final answer contains only the selected tool name. Accuracy is computed by normalized exact match against the reference tool name. GPQA-diamond (198 examples) provides a decode-heavy contrast and measures exact-match accuracy of the final answer option following longer generations~\citep{rein2023gpqagraduatelevelgoogleproofqa}. Across all three tasks, we evaluate paired Instruct and Thinking regimes to examine how inference regime changes task quality and online serving cost.

\subsection{Models, variants, and serving strategies}

We evaluate Qwen3 models under Instruct and Thinking regimes, using Qwen3-8B for Moderation and Tool-use and Qwen3-30B-A3B for the main GPQA experiments~\citep{yang2025qwen3technicalreport}. A supplementary GPQA control uses Qwen3-8B in Thinking mode. We additionally evaluate Claude, DeepSeek, and GLM on Moderation to assess whether the quality gains extend across model families. The cross-task cost comparisons use the stated Qwen configurations. Model versions and decoding details are provided in Appendix~\ref{sec:appendix-implementation}.

We compare five main experience settings: \textit{None}, \textit{Random experience}, \textit{Global compact}, \textit{Global full}, and \textit{Retrieval}. In GPQA, we additionally evaluate \textit{Trigger-aware} retrieval and an \textit{LLM selector}. These settings separate no-experience baselines, unmatched-context controls, static always-on serving baselines, and case-dependent selective serving. Trigger-aware examines query-construction overhead; the LLM selector is a high-overhead headroom estimate. Retrieval budgets are selected on an independent development set, not on the final test sets.

For moderation, all variants share the same category taxonomy in the system prompt. The experience repository is built offline from business-labeled training data, while the evaluation set is disjoint and deduplicated so that experience items provide reusable case-level operational knowledge beyond the shared taxonomy rather than redefining the label space.

For Tool-use, experience is constructed from historical and operational tool-routing materials separate from the evaluation workload. GPQA experience is extracted from non-Diamond GPQA questions, with all 198 Diamond evaluation questions excluded before extraction. Repository sources and evaluation-isolation details are provided in Appendix~\ref{sec:appendix-experience-unit}.

\subsection{Metrics and quality--cost perspective}
\label{sec:metrics}

We evaluate both task quality and online serving cost. For Moderation, the primary metric is overall accuracy: a prediction is correct only when its complete classification path matches the gold path, including the terminal leaf. This metric is evaluated over all 1,500 examples.

We additionally report Macro-F1 and hierarchical path accuracies. L1 path accuracy evaluates all 1,500 examples; L1+L2 path accuracy evaluates the 912 examples whose gold leaves are at L2 or L3; and L1+L2+L3 path accuracy evaluates the 237 examples whose gold leaves are at L3. Each path metric requires joint correctness of all listed levels. The evaluation subsets are determined by gold annotation depth and remain identical across serving strategies.

Benign FPR measures the fraction of gold-benign inputs assigned any non-benign risk category and provides a complementary measure of over-enforcement. Because the evaluation class proportions do not mirror live traffic, overall accuracy is used for controlled comparisons among serving strategies rather than as an estimate of online accuracy.

We define \textit{full request-path latency} (\textit{Full E2E}) as the client-observed wall-clock time from request initiation to receipt of the complete response. For Retrieval, this includes query construction, candidate retrieval and fusion, reranking, prompt assembly, and model inference (prefill and decoding). The same timing boundary applies to all serving strategies, including the stages performed by each strategy. Any additional online selector belongs to this complete path; standalone selector-stage timings are reported separately and are not treated as Full E2E.

Prompt and completion tokens provide complementary indicators of prompt-side and generation-side workload. They help interpret latency differences but do not provide a complete prefill/decode runtime decomposition. Prefix caching is enabled unless a configuration is explicitly identified as cache-off. Hardware and measurement details are provided in Appendix~\ref{sec:appendix-serving-protocol}.

Unless otherwise stated, reported metrics for Moderation and Tool-use are averaged over three runs, whereas GPQA-diamond metrics are averaged over eight runs; standard deviations, where reported, are computed across the corresponding runs. Repeated runs assess stability on a fixed test set rather than variation across independent test sets. Paired bootstrap confidence intervals and McNemar tests are reported in Appendix Table~\ref{tab:appendix-statistics}.

Appendix~\ref{sec:appendix-shortoutput-throughput} reports additional short-output measurements of request throughput (QPS).

\section{Results}

\subsection{Serving-interface comparison: selective retrieval vs. unconditional injection}

\begin{table}[t]
\centering
\small
\setlength{\tabcolsep}{2.8pt}
\begin{tabular}{llrr}
\toprule
Task & Setting & Acc. (\%) & Full E2E (s) \\
\midrule
\multirow{5}{*}{Moderation}
& None & $46.7\pm0.2$ & 0.29 \\
& Random & $57.0\pm0.4$ & 0.29 \\
& Global compact & $69.4\pm0.3$ & 0.30 \\
& Global full 50 & $63.1\pm0.2$ & 0.41 \\
& Retrieval K10 & $82.1\pm0.1$ & 0.58 \\
\midrule
\multirow{5}{*}{Tool-use}
& None & $67.9\pm0.1$ & 0.08 \\
& Random & $69.5\pm0.1$ & 0.09 \\
& Global compact & $71.9\pm0.2$ & 0.12 \\
& Global full 50 & $72.9\pm0.2$ & 0.25 \\
& Retrieval K5 & $89.0\pm0.1$ & 0.56 \\
\midrule
\multirow{5}{*}{GPQA}
& None & $61.9\pm0.2$ & 19.35 \\
& Random & $62.0\pm0.3$ & 23.16 \\
& Global compact & $62.8\pm0.5$ & 25.08 \\
& Global full 50 & $61.5\pm0.2$ & 33.26 \\
& Retrieval K10 & $64.7\pm0.4$ & 16.92 \\
& Trigger-aware & $66.7\pm0.3$ & 17.58 \\
& LLM selector & $78.0\pm0.5$ & 76.62 \\
\bottomrule
\end{tabular}
\caption{Instruct accuracy (mean $\pm$ SD; three runs for Moderation/Tool-use, eight for GPQA) and Full E2E. Token counts: Appendix Table~\ref{tab:appendix-interface-details}.}
\label{tab:main-comparison}
\end{table}

Table~\ref{tab:main-comparison} compares accuracy and Full E2E in the Instruct regime. On Moderation, Retrieval achieves 82.1\% accuracy, with gains of 35.4 percentage points (pp) over None and 12.7~pp over Global compact. Random reaches 57.0\%, leaving a gap of 25.1~pp to Retrieval. Global full 50 reaches 63.1\% despite using a much larger prompt than Retrieval (18.1K versus 3.31K tokens). Request-conditioned retrieval thus yields higher accuracy than both random experience and fixed global injection.

On Tool-use, Retrieval K5 achieves 89.0\% tool-selection accuracy, exceeding the strongest static baseline, Global full 50, by 16.1~pp and None by 21.1~pp.

On GPQA, Retrieval reaches 64.7\% accuracy at 16.92\,s, compared with 61.9\% at 19.35\,s for None. Random and the global-injection baselines have Full E2E values ranging from 23.16\,s to 33.26\,s.

Across the short-output workloads, the accuracy gains accompany higher Full E2E; GPQA instead shows higher observed accuracy alongside lower latency. We next examine the classification gains and the serving conditions behind these cost patterns.

\begin{figure*}[t]
\centering
\includegraphics[width=\linewidth]{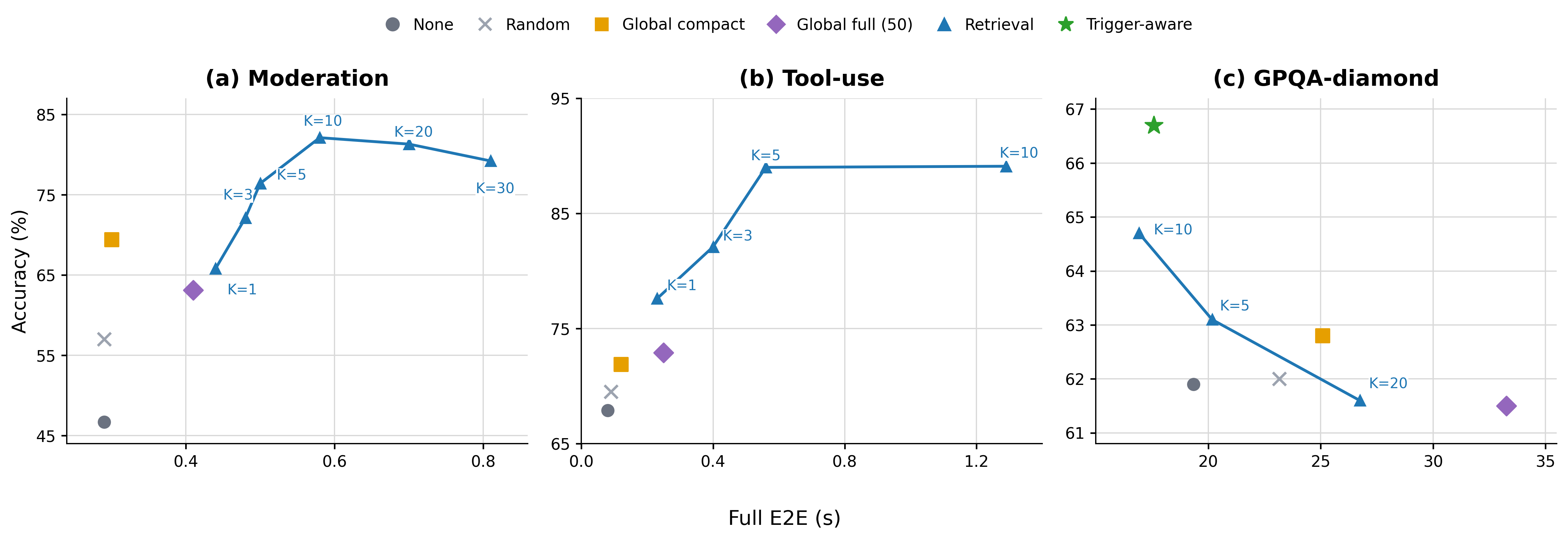}
\caption{Instruct quality--latency operating points with task-specific axes. Lines connect Retrieval settings in increasing $K$. Moderation sweep and cache timings are measured separately.}
\label{fig:quality-cost}
\end{figure*}

\subsection{Understanding quality gains: fine-grained moderation and cross-model checks}

\begin{table}[t]
\centering
\small
\setlength{\tabcolsep}{3pt}
\begin{tabular}{@{}lrrr@{}}
\toprule
Metric (\%) & None & Global compact & Retrieval \\
\midrule
Overall accuracy & 46.7 & 69.4 & \textbf{82.1} \\
Macro-F1 & 8.1 & 27.1 & \textbf{50.2} \\
\addlinespace[2pt]
L1 path accuracy & 79.8 & 83.3 & \textbf{89.8} \\
L1+L2 path accuracy & 30.5 & 57.1 & \textbf{77.3} \\
L1+L2+L3 path accuracy & 7.5 & 56.5 & \textbf{76.4} \\
\addlinespace[2pt]
Benign FPR $\downarrow$ & 6.4 & \textbf{5.2} & 7.1 \\
\bottomrule
\end{tabular}
\caption{Moderation diagnostics (Qwen3-8B Instruct). Overall accuracy requires exact full-path matches on all 1,500 examples. L1, L1+L2, and L1+L2+L3 use 1,500, 912, and 237 examples, respectively, with gold leaves at or beyond the indicated depth. Path metrics require joint correctness. Benign FPR uses 566 benign examples. Bold marks the best mean. Supplementary metrics and uncertainties: Appendix Table~\ref{tab:appendix-moderation-risk}.}
\label{tab:moderation-quality-diagnostics}
\end{table}

\textbf{Fine-grained operational distinctions.}
All Moderation variants share the same taxonomy with 16/64/94 categories across three levels. Table~\ref{tab:moderation-quality-diagnostics} shows that Retrieval increases overall accuracy from 46.7\% to 82.1\% and macro-F1 from 8.1\% to 50.2\%. On the 237 examples whose gold leaves are at L3, complete three-level path accuracy increases from 7.5\% to 76.4\%. Path accuracy also improves at the other evaluated depths, with each comparison using the same gold-defined subset across serving strategies. These results demonstrate improvements in both overall classification and fine-grained hierarchical decisions.

The qualitative examples in Appendix~\ref{sec:appendix-case-analysis} illustrate how retrieved rules can supply applicability conditions, policy-conditioned distinctions, and adjacent-category boundary cues.

The overall accuracy gain reflects improved classification of risk examples, while benign FPR increases from 6.4\% to 7.1\% (Appendix Table~\ref{tab:appendix-moderation-risk}). Global compact has a lower benign FPR of 5.2\%, providing an alternative operating point when benign over-enforcement is a stronger deployment constraint.

\begin{table}[t]
\centering
\small
\setlength{\tabcolsep}{4pt}
\resizebox{\linewidth}{!}{%
\begin{tabular}{lrrr}
\toprule
Model & None & Global compact & Retrieval K10 \\
\midrule
Claude-Sonnet-5.5 & 67.6 $\pm$ 0.5 & 74.5 $\pm$ 0.7 & 90.4 $\pm$ 0.3 \\
DeepSeek-v4-flash & 69.3 $\pm$ 0.9 & 73.8 $\pm$ 0.5 & 86.0 $\pm$ 0.4 \\
GLM-5.3-flash & 62.5 $\pm$ 0.8 & 72.5 $\pm$ 0.4 & 86.1 $\pm$ 0.3 \\
\bottomrule
\end{tabular}
}
\caption{Moderation accuracy (\%) across additional model families, reported as the three-run mean $\pm$ standard deviation.}
\label{tab:cross-model-moderation}
\end{table}

\textbf{Cross-model checks.}
Retrieval also improves Moderation accuracy across Claude, DeepSeek, and GLM (Table~\ref{tab:cross-model-moderation}). Its gains over Global compact range from 12.2~pp to 15.9~pp, reproducing the benefit of request-conditioned experience across these model families.

\subsection{Quality--cost trade-offs across workloads and serving budgets}

\begin{table*}[t]
\centering
\small
\setlength{\tabcolsep}{4pt}
\begin{tabular}{llrrrr}
\toprule
Task & Inference mode & Accuracy (\%) & Prompt (K)
     & Completion tokens & Full E2E (s) \\
\midrule
\multirow{2}{*}{Moderation}
& Instruct
& $46.7 \rightarrow 82.1$
& $1.26 \rightarrow 3.31$
& $28.3 \rightarrow 30.5$
& $0.29 \rightarrow 0.58$ \\
& Thinking
& $58.1 \rightarrow 85.6$
& $1.26 \rightarrow 3.31$
& $401.8 \rightarrow 475.0$
& $3.08 \rightarrow 3.92$ \\
\midrule
\multirow{2}{*}{Tool-use}
& Instruct
& $67.9 \rightarrow 89.0$
& $0.52 \rightarrow 4.75$
& $2.8 \rightarrow 2.6$
& $0.08 \rightarrow 0.56$ \\
& Thinking
& $77.9 \rightarrow 92.0$
& $0.52 \rightarrow 4.74$
& $288.1 \rightarrow 372.5$
& $4.40 \rightarrow 8.73$ \\
\midrule
\multirow{2}{*}{GPQA}
& Instruct
& $61.9 \rightarrow 64.7$
& $0.27 \rightarrow 1.85$
& $2.44\mathrm{K} \rightarrow 1.42\mathrm{K}$
& $19.35 \rightarrow 16.92$ \\
& Thinking
& $68.3 \rightarrow 70.0$
& $0.25 \rightarrow 1.91$
& $6.31\mathrm{K} \rightarrow 5.12\mathrm{K}$
& $72.55 \rightarrow 63.09$ \\
\bottomrule
\end{tabular}
\caption{Mean values for None $\rightarrow$ Retrieval across inference modes ($K=10$ for Moderation/GPQA; $K=5$ for Tool-use). Details: Appendix Tables~\ref{tab:appendix-regime-matrix} and~\ref{tab:appendix-statistics}.}
\label{tab:breakeven}
\end{table*}

\textbf{Output structure and inference regime.}
Retrieval increases Full E2E on Moderation and Tool-use in both inference modes (Table~\ref{tab:breakeven}). Completion length changes little in Instruct and increases in Thinking. On GPQA, Retrieval reduces Full E2E by 12.6\% in Instruct and 13.0\% in Thinking, with shorter completions despite larger prompts. This pattern is consistent with downstream generation savings offsetting added prompt-side work. The Qwen3-8B Thinking GPQA control also shows shorter completions and lower Full E2E (Appendix Table~\ref{tab:appendix-gpqa-8b-control}).

\textbf{Retrieval budget.}
In the Moderation sweep, accuracy rises from 65.8\% at $K=1$ to its highest observed value of 82.1\% at $K=10$ (Figure~\ref{fig:quality-cost}). Increasing $K$ to 30 lowers accuracy to 79.2\% and raises Full E2E from 0.58\,s to 0.81\,s. In Tool-use, accuracy increases across the plotted $K=1$, $K=3$, and $K=5$ configurations. Increasing $K$ from 5 to 10 then changes Instruct accuracy from 89.0\% to 89.1\%, while Full E2E rises from 0.56\,s to 1.29\,s. Under Thinking, the larger Tool-use budget yields lower observed accuracy and higher latency. On GPQA, the measured $K=10$ configuration has higher observed accuracy and lower Full E2E than $K=5$ and $K=20$, illustrating that the latency response to retrieval depth can be non-monotonic. These operating points show that the value of increasing $K$ depends on the task and current budget. Tabulated budget measurements appear in Appendix Table~\ref{tab:appendix-budget-points}, with the Tool-use inference-mode comparison in Appendix Table~\ref{tab:appendix-tooluse-topk}.

\textbf{Prefix-cache reuse.}
In the separate cache benchmark, enabling prefix caching reverses the latency ordering of Global compact and Retrieval $K=10$. With caching off, their Full E2E values are 4.90\,s and 1.49\,s; with caching on, they become 0.30\,s and 0.58\,s. Global compact has a reported prefix hit rate of 99.7\%, compared with 39.9\% for Retrieval. Under the measured cache-on load, Global compact also achieves higher request throughput (65.86 versus 34.12 QPS). These results show how prefix reuse changes the serving value of static and request-conditioned experience. Full cache and throughput measurements appear in Appendix Tables~\ref{tab:appendix-prefix-cache} and~\ref{tab:appendix-shortoutput-throughput}.

\textbf{Selection overhead.}
GPQA illustrates the cost of more expensive experience selection. Trigger-aware retrieval reaches 66.7\% accuracy versus 64.7\% for standard Retrieval, with a full request-path E2E of 17.58\,s, including 0.76\,s of additional model-based selection overhead. The LLM selector reaches 78.0\% at a measured Full E2E of 76.62\,s, including 58.81\,s of selector-stage latency; selection uses 55.35K tokens (Appendix~\ref{sec:appendix-selector-overhead}).

\textbf{Configuration choice.}
The measured accuracy--latency points distinguish configurations with advantages on both metrics from those requiring a trade-off. In Moderation, Global compact offers higher accuracy and lower Full E2E than Global full 50; Retrieval K10 likewise improves both over K20 and K30. Tool-use K10 instead gains 0.1~pp over K5 in Instruct while more than doubling Full E2E. On GPQA, standard Retrieval improves both metrics over None, Random, and global injection, while Trigger-aware trades higher observed accuracy for higher latency. These comparisons use observed means; deployment choices also depend on task-specific false-positive and service constraints.

\section{Discussion}

\textbf{Match serving choices to operational requirements.}
Our deployments use Retrieval $K=10$ for Moderation, Retrieval $K=5$ for Tool-use, and Trigger-aware retrieval for GPQA-like reasoning. The latter two operate within agent workflows with less stringent latency requirements. Moderation illustrates why quality and service constraints must be considered together: retrieval improves fine-grained categorization but increases benign FPR, while Global compact offers lower over-enforcement and cache-on latency. The latency reversal between these interfaces under prefix caching shows that uncached comparisons can favor a different configuration. Deployment assessment should therefore combine application-specific quality, false-positive, and latency requirements under the intended cache policy and load.

\textbf{Account for the downstream effects of experience budgets.}
An experience budget affects retrieval and selection, prompt processing, and subsequent generation. Short-output workloads offer limited generation time to save. On GPQA, longer prompts accompany shorter completions and lower Full E2E, consistent with generation savings offsetting additional serving work, although our measurements do not isolate these components. Tool-use's quality plateau beyond $K=5$ and the LLM selector's substantial overhead illustrate diminishing returns and expensive quality gains, respectively. Budget selection should compare incremental task quality with complete-path latency on independent development data under the target serving conditions; prompt and completion lengths help interpret the trade-off.

\section{Conclusion}

We studied how external experience serving changes task quality and deployment choices. Retrieval improves fine-grained Moderation quality across the evaluated model families, while Global compact retains advantages in cache-on latency, request throughput, and benign FPR. Prefix caching can reverse the latency ordering of these interfaces; in GPQA, shorter completions coincide with lower Full E2E despite larger prompts. Budget and selector comparisons further distinguish improvements in both observed quality and latency from quality gains obtained at additional serving cost. These findings support selecting experience-serving configurations under explicit quality and service constraints, using complete-path measurements that account for retrieval, selection, and generation.

\clearpage

\section*{Limitations}

This paper studies \textit{external experience serving} at inference time rather than the full lifecycle of experience learning, including repository construction, maintenance, or internalization into model parameters.

We measure task quality and Full E2E, using prompt and completion tokens to help interpret cost differences. These measurements do not isolate the runtime contributions of retrieval, prefill, and decoding. Cache and throughput results are specific to the measured serving configurations, and do not establish a general monetary-cost or production-capacity model.

The Moderation benchmark is risk-oriented rather than traffic-matched, so aggregate accuracy supports controlled strategy comparisons rather than estimation of live-traffic utility. Overall benign FPR does not replace sensitive-slice evaluation. Additional model families replicate the Moderation quality direction, while the cross-task cost evidence remains tied to the stated Qwen configurations. The same-generator GPQA control reduces the model-size confound but does not isolate task and repository effects. GPQA experience construction and evaluation use disjoint questions from the same benchmark; this comparison does not establish transfer to other scientific QA datasets.

The LLM selector has a measured Full E2E of 76.62\,s and estimates quality headroom at substantial selection cost. Its selection tokens are reported separately from downstream generation tokens. Some managed-service stage-level parameters and proprietary operational details remain undisclosed or unavailable, limiting component-level reproducibility.

\section*{Ethical Considerations}

This work studies external experience serving using production-derived moderation data and business-labeled examples. To reduce privacy and exposure risks, we do not disclose raw production content, private identifiers, or verbatim user traces. The paper instead reports aggregate statistics, abstracted prompt schemas, and sanitized experience-item formats. The evaluation set is separated from the repository source data, and deduplication is applied to reduce overlap between evaluation instances and repository items.

Moderation errors can cause harm through both under-enforcement and over-enforcement. False negatives may allow unsafe content to pass, while false positives may incorrectly penalize benign, ambiguous, or context-sensitive expression. Our qualitative examples illustrate how retrieved rules can over-trigger on benign or context-sensitive inputs, leading to inappropriate risk labels. Deployment should therefore include evaluation on relevant benign and sensitive slices, application-specific balancing of false positives and false negatives, and human review of uncertain or potentially high-impact decisions where appropriate.



\bibliography{custom}

\appendix

\section{Additional implementation details}
\label{sec:appendix-implementation}

This appendix provides additional implementation details for the experience-serving variants used in the main paper, along with supplementary comparisons referenced in the main text.

\subsection{Task and evaluation details}
\label{sec:appendix-task-metrics}

Table~\ref{tab:appendix-task-metrics} summarizes the task definition and primary evaluation metric for each task family. Unless otherwise noted, metrics for Moderation and Tool-use are averaged over three runs, whereas GPQA-diamond metrics are averaged over eight runs.

\begin{table*}[t]
\centering
\small
\begin{tabular}{p{2.1cm}p{4.8cm}p{7.6cm}}
\toprule
Task & Input / output & Evaluation \\
\midrule
Moderation & User query or content to be moderated $\rightarrow$ hierarchical safety label & Overall accuracy (\%) requires an exact match of the complete path ending at the gold leaf, evaluated over all 1,500 examples (934 risk and 566 benign). Gold leaves may be at L1, L2, or L3; benign is an L1 leaf. \\
Tool-use & User request and tool descriptions $\rightarrow$ selected tool name & Accuracy (\%) is computed by normalized exact match against the reference tool name. \\
GPQA-diamond & Multiple-choice question $\rightarrow$ A/B/C/D & Exact-match accuracy (\%) on the final answer option. \\
\bottomrule
\end{tabular}
\caption{Task and metric definitions used in the appendix and main results.}
\label{tab:appendix-task-metrics}
\end{table*}

\subsection{Moderation task protocol}
\label{sec:appendix-moderation-protocol}

Our primary moderation task is derived from a real production LLM safety-guard workload rather than from a synthetic benchmark. The task is formulated as hierarchical multi-class safety categorization with 16 top-level categories, 64 second-level categories, and 94 third-level categories. The top- and second-level taxonomy follows the TC260 standard for generative-AI safety categories, while the finer-grained hierarchy is used internally for operational labeling.

The Moderation benchmark reported in this paper contains 1,500 examples constructed for industrial risk-oriented assessment rather than natural traffic-rate estimation: 934 risk examples and 566 benign examples. The risk/benign mixture is intentionally not a natural-traffic prevalence estimate. Annotation begins with LLM-generated preliminary labels. Three human annotators then perform cross-annotation, followed by verification by a quality inspector. Annotation-consistency checks exceed 95\%. For both the Instruct and Thinking regimes, the category taxonomy is included in the system prompt as shared task specification for all moderation variants.

Each Moderation example has a gold terminal leaf at L1, L2, or L3; benign is an L1 leaf. Overall accuracy requires an exact match of the complete path from L1 to the gold leaf and is evaluated over all 1,500 examples. The hierarchical diagnostics use nested gold-defined subsets: L1 includes all 1,500 examples, L1+L2 includes the 912 examples whose gold leaves are at L2 or L3, and L1+L2+L3 includes the 237 examples whose gold leaves are at L3. Each diagnostic requires all listed levels to match simultaneously, using the same subset across serving strategies.

The moderation experience repository and the benchmark evaluation set are separated at the data level, and deduplication is applied to reduce overlap between repository items and evaluation instances, including the adversarially augmented risk cases.

\begin{table*}[t]
\centering
\small
\setlength{\tabcolsep}{3.5pt}
\renewcommand{\arraystretch}{1.08}
\textbf{(a) Classification and hierarchical-path quality (\%)}\par\vspace{3pt}
\begin{tabular*}{\linewidth}{@{\extracolsep{\fill}}lrrrrr@{}}
\toprule
Setting & Overall Acc. & Macro-F1 & L1 & L1+L2 & L1+L2+L3 \\
\midrule
None & 46.7 $\pm$ 0.2 & 8.1 $\pm$ 0.3 & 79.8 & 30.5 & 7.5 \\
Retrieval & 82.1 $\pm$ 0.1 & 50.2 $\pm$ 1.2 & 89.8 & 77.3 & 76.4 \\
Global compact & 69.4 $\pm$ 0.3 & 27.1 $\pm$ 0.8 & 83.3 & 57.1 & 56.5 \\
Random & 57.0 $\pm$ 0.4 & 14.3 $\pm$ 0.6 & 78.0 & 40.7 & 27.7 \\
\bottomrule
\end{tabular*}
\par\vspace{6pt}
\textbf{(b) Risk and benign false-positive diagnostics (\%)}\par\vspace{3pt}
\begin{tabular*}{\linewidth}{@{\extracolsep{\fill}}lrrrrr@{}}
\toprule
Setting & Risk recall & Risk FNR & Benign FPR & Specificity & Balanced Acc. \\
\midrule
None & 95.7 $\pm$ 0.1 & 4.3 $\pm$ 0.1 & 6.4 $\pm$ 0.0 & 93.6 $\pm$ 0.0 & 94.7 $\pm$ 0.1 \\
Retrieval & 98.8 $\pm$ 0.1 & 1.2 $\pm$ 0.1 & 7.1 $\pm$ 0.2 & 92.9 $\pm$ 0.2 & 95.9 $\pm$ 0.1 \\
Global compact & 94.7 $\pm$ 0.1 & 5.3 $\pm$ 0.1 & 5.2 $\pm$ 0.1 & 94.8 $\pm$ 0.1 & 94.7 $\pm$ 0.1 \\
Random & 90.9 $\pm$ 0.3 & 9.1 $\pm$ 0.3 & 4.4 $\pm$ 0.1 & 95.6 $\pm$ 0.1 & 93.3 $\pm$ 0.1 \\
\bottomrule
\end{tabular*}
\caption{Moderation results on 1,500 examples: 934 risk and 566 benign. Overall Acc. measures exact matches of complete paths to the gold leaves. L1 uses all 1,500 examples; L1+L2 uses 912 examples with gold leaves at depth two or greater; L1+L2+L3 uses 237 examples with gold leaves at depth three. ``+'' denotes joint correctness of the listed levels. Risk recall and FNR use gold-risk examples; benign FPR and specificity use gold-benign examples.}
\label{tab:appendix-moderation-risk}
\end{table*}

\paragraph{Risk metric interpretation.}
Risk recall is computed on gold-risk examples, and risk FNR is its complement. Benign FPR is computed on gold-benign examples, and specificity is its complement. The binary risk diagnostics are supplementary to overall accuracy and Macro-F1; the risk/benign mixture in this evaluation set is not a natural-traffic prevalence estimate.

\begin{table}[t]
\centering
\small
\begin{tabular}{llr}
\toprule
Model & Setting & Accuracy (\%) \\
\midrule
Claude-Sonnet-5.5 & None & 67.60 $\pm$ 0.46 \\
Claude-Sonnet-5.5 & Random & 68.40 $\pm$ 0.41 \\
Claude-Sonnet-5.5 & Global compact & 74.50 $\pm$ 0.66 \\
Claude-Sonnet-5.5 & Retrieval K10 & 90.40 $\pm$ 0.26 \\
DeepSeek-v4-flash & None & 69.34 $\pm$ 0.86 \\
DeepSeek-v4-flash & Random & 70.12 $\pm$ 0.68 \\
DeepSeek-v4-flash & Global compact & 73.80  $\pm$ 0.52 \\
DeepSeek-v4-flash & Retrieval K10 & 85.95 $\pm$ 0.40 \\
GLM-5.3-flash & None & 62.47 $\pm$ 0.81 \\
GLM-5.3-flash & Random & 65.67 $\pm$ 0.25 \\
GLM-5.3-flash & Global compact & 72.53 $\pm$ 0.44 \\
GLM-5.3-flash & Retrieval K10 & 86.13 $\pm$ 0.25 \\
\bottomrule
\end{tabular}
\caption{Cross-model Moderation replication on the evaluation set.}
\label{tab:appendix-cross-model-moderation}
\end{table}

\begin{table*}[t]
\centering
\small
\begin{tabular}{lllll}
\toprule
Task & Comparison & Delta Accuracy & 95\% CI & McNemar $p$ (min--max) \\
\midrule
Moderation & Retrieval K10 $-$ None & +35.4 pp & [32.76, 37.98] & $[2.17\times10^{-126},\,3.91\times10^{-123}]$ \\
Moderation & Retrieval K10 $-$ Random & +25.1 pp & [22.49, 27.65] & $[1.34\times10^{-73},\,3.93\times10^{-70}]$ \\
Moderation & Retrieval K10 $-$ Global compact & +12.7 pp & [10.31, 15.18] & $[3.90\times10^{-24},\,3.21\times10^{-22}]$ \\
Moderation & K20 $-$ K10 & $-$0.8 pp & [$-$1.75, 0.15] & $[0.09 ,0.30]$ \\
Moderation & K30 $-$ K10 & $-$2.9 pp & [$-$4.97, $-$0.90] & [0.007 ,0.017] \\
Tool-use & Retrieval K5 $-$ None & +21.12 pp & [17.84, 24.39] & $[1\times10^{-41},\,1\times10^{-24}]$ \\
Tool-use & Retrieval K10 $-$ K5 & +0.12 pp & [$-$1.37, 1.62] & [0.996, 1.000]\\
GPQA & Retrieval $-$ None & +2.78 pp & [$-$5.59, 11.65] & [0.41 ,0.63] \\
GPQA & Trigger-aware $-$ standard & +1.96 pp & [$-$6.26, 10.30] & [0.61 ,0.77] \\
\bottomrule
\end{tabular}
\caption{Paired uncertainty summaries}
\parbox{\linewidth}{\footnotesize

All reported 95\% CIs are computed using 10,000 paired bootstrap resamples of evaluation items. Each resample sampled item IDs with replacement while retaining the paired predictions from both conditions. Accuracy differences are reported in percentage points. McNemar \(p\)-values use
the stated two-sided test. A \(p\)-value range denotes the
minimum--maximum value across runs. All \(p\)-values are unadjusted; no
multiple-comparison correction is applied.

}
\label{tab:appendix-statistics}
\end{table*}

\subsection{Experience unit and repository construction}
\label{sec:appendix-experience-unit}

Throughout the paper, an \textit{experience item} denotes one reusable unit of externalized operational knowledge. Depending on the task, an experience item may correspond to a policy rule, a prior failure case, a resolved edge case, a tool-use pattern, or a short decision trace that can be reused at inference time.

Our study focuses on \textit{serving} existing repositories rather than constructing them. The experience collections are prepared offline from task-specific source materials, separately from the corresponding final evaluation sets.

For Moderation, the experience repository is prepared from approximately 300K business-labeled training examples using an ACE-style extraction pipeline~\citep{zhang2026agenticcontextengineeringevolving}, yielding over 105K experience items. The evaluation set is disjoint from the repository source data, and deduplication is applied to reduce overlap. The extracted items provide reusable operational cues beyond the shared category definitions, such as prior edge cases and category-boundary patterns.

For Tool-use, the 800-example evaluation workload is derived from production tool-routing scenarios. Each example contains a user request, descriptions of the available tools, and a reference tool name. The repository contains 465 experience items constructed from separate historical and operational routing materials.

For GPQA-diamond, the repository contains 559 experience items extracted from the \texttt{gpqa\_experts}, \texttt{gpqa\_main}, and \texttt{gpqa\_extended} configurations of \texttt{Idavidrein/gpqa}~\citep{rein2023gpqagraduatelevelgoogleproofqa}. All 198 GPQA-diamond evaluation questions are excluded from these source configurations before experience extraction. Experience construction and final evaluation therefore use disjoint question sets from the same benchmark.

Across all three tasks, repository construction excludes final evaluation inputs and their reference annotations, model outputs generated for those inputs, and evaluation-specific reflections.

\paragraph{Experience item format.}
We do not disclose raw production content in the appendix. Instead, each experience item is stored as a short, abstracted, and reusable guidance unit. A typical item contains a compact applicability trigger, a recurring failure pattern, an actionable correction or decision rule, and optionally a sanitized abstract example. This format allows the repository to preserve reusable operational knowledge while avoiding raw user inputs, private identifiers, or verbatim production traces.

\begin{quote}
\small
\texttt{[bullet\_id]}\\[2pt]
\\
\texttt{[Trigger]}\\ A short description of when this experience may apply.\\[2pt]
\\
\texttt{[Common Failure]}\\ A recurring model mistake observed in previous cases.\\[2pt]
\\
\texttt{[Action]}\\ The expected correction or decision rule.\\[2pt]
\\
\texttt{[Optional Example]}\\ A sanitized or abstracted example.
\end{quote}

\subsection{Model and decoding settings}
\label{sec:appendix-model-settings}

For moderation and tool-use, we use Qwen3-8B under paired Instruct and Thinking regimes, realized through \texttt{no\_think} / \texttt{think} prompt control. For the main GPQA-diamond experiments, we use Qwen3-30B-A3B-Instruct-2507 and Qwen3-30B-A3B-Thinking-2507 as the corresponding Instruct and Thinking models. The supplementary GPQA control uses Qwen3-8B in Thinking mode, matching the generator used for Moderation and Tool-use (Table~\ref{tab:appendix-gpqa-8b-control}). Additional Moderation quality evaluations use Claude-Sonnet-5.5, DeepSeek-v4-flash, and GLM-5.3-flash. We follow the model-card recommended decoding settings for the respective Qwen regimes.

Specifically, for the Thinking regime we use \texttt{Temperature=0.6}, \texttt{TopP=0.95}, and \texttt{TopK=20}. For the Instruct regime, we use \texttt{Temperature=0.7}, \texttt{TopP=0.8}, and \texttt{TopK=20}. These decoding settings are shared across moderation, tool-use, and GPQA; the main difference across tasks lies in the model family and the task-specific experience repository.

\paragraph{Serving setup.}
We serve the generation models with vLLM and assemble the final prompt only after the retrieval or selection step is completed. Unless otherwise stated, the main serving experiments are run with prefix caching enabled. The appendix reports Full E2E, prompt tokens, and completion tokens, with QPS additionally reported for the short-output serving comparison. For retrieval-based settings, Full E2E includes query construction, candidate retrieval and reranking, prompt assembly, and downstream generation. For Trigger-aware and the LLM selector, it also includes model-based selection; selector-stage latency and tokens are additionally reported as component measurements.

\subsection{Serving hardware and measurement protocol}
\label{sec:appendix-serving-protocol}

\paragraph{Hardware and model configuration.}
For the locally hosted experiments, we use the official Hugging Face Qwen checkpoints: Qwen3-8B for Moderation and Tool-use and Qwen3-30B-A3B-Instruct/Thinking-2507 for GPQA-diamond, without quantization. The inference server is equipped with eight NVIDIA A800 GPUs, each with 80\,GB of memory, and runs vLLM v0.16.0. Prefix caching is enabled except in explicitly marked cache-off conditions. The serving configuration sets a maximum sequence length of 131,072 tokens and uses YaRN RoPE scaling with factor 4.0 and an original context length of 32,768 tokens. The relevant launch options are:
\begin{quote}
\small
\begin{verbatim}
--rope-scaling
'{"rope_type":"yarn","factor":4.0,
"original_max_position_embeddings":
32768}'
--max-model-len 131072
\end{verbatim}
\end{quote}
The excerpt specifies the RoPE configuration as a JSON argument to \texttt{vllm serve}. For managed-service models, we report Moderation accuracy only.

\paragraph{Request workload and concurrency.}
Each main serving condition uses 20 concurrent requests and receives the complete fixed evaluation workload for its task: 1,500 Moderation examples, 800 Tool-use examples, or 198 GPQA-diamond examples, respectively. Each example is submitted once per run. Within a run, the request contents and execution order are kept consistent across serving conditions. The cache-aware serving and throughput measurements use the Moderation workload under the same concurrency setting.

\paragraph{Warm-up, cache reset, and condition order.}
Before timed measurement, the model server is fully initialized and a warm-up phase is performed; warm-up requests are excluded from the reported measurements. Cache-on and cache-off measurements are prepared as separate server states, and the cache state is reset or reinitialized when switching between the corresponding conditions so that cached state is not carried across measurements. The serving conditions are evaluated in a consistent order using the same request workload for each repeated measurement.

\paragraph{Latency measurement and aggregation.}
For each condition, Full E2E is computed as the arithmetic mean of the latency of all measured requests, from client-side request initiation until the complete response is received. Thus, the reported Full E2E values are request-level means rather than medians or selected percentiles. Moderation and Tool-use use three independent runs, whereas GPQA-diamond uses eight runs. Repeated runs use the same fixed evaluation workloads and are used to assess run-to-run stability. The dedicated cache-aware comparison additionally reports P95 as a separate tail-latency statistic.

\subsection{Prompt schema and injection formats}
\label{sec:appendix-prompt-schema}

Across all settings, the prompt follows the same high-level structure: a task-specific system instruction with shared taxonomy or task guidance, an optional experience block, the current user input or task instance, and a constrained output format. We provide an abstracted prompt schema with task-specific and production-sensitive content represented by placeholders.

\paragraph{General prompt schema.}
The following placeholder schema summarizes the shared prompt structure used across tasks:

\begin{quote}
\small
\textbf{System:}\\
You are a task-specific assistant for \{TASK\_NAME\}. Follow the taxonomy and task instructions below.\\[2pt]
\\
\texttt{[Taxonomy]} \\ \{TASK\_TAXONOMY\_PLACEHOLDER\}\\
\\
\texttt{[Task Instruction]} \\ \{TASK\_INSTRUCTION\_PLACEHOLDER\}\\
\\
\texttt{[Optional Experience Block]} \\ \{EXPERIENCE\_BLOCK\_PLACEHOLDER\}\\[2pt]
\\
\textbf{User:}\\
\{USER\_INPUT\_PLACEHOLDER\}\\[2pt]
\\
\textbf{Output Format:}\\
\{OUTPUT\_FORMAT\_PLACEHOLDER\}
\end{quote}

\paragraph{No-experience and retrieval-based prompts.}
The no-experience baseline uses the task instruction and output format without an experience block. Retrieval-based injection uses the same prompt scaffold, but fills the optional experience block with the Top-$K$ retrieved experience items for the current input. Retrieved items are inserted as a structured list of short reusable guidance bullets rather than as free-form long context.

\begin{quote}
\small
\texttt{[Retrieved Experience]}\\
1. \{retrieved\_experience\_1\}\\
2. \{retrieved\_experience\_2\}\\
...\\
K. \{retrieved\_experience\_K\}\\[2pt]
\\
\texttt{[Instruction]}\\ Use the retrieved experience only when it is relevant. Do not force an experience if it does not apply.
\end{quote}

\paragraph{Random, global full, and global compact variants.}
For the random baseline, we randomly sample experience items from the same task-specific repository without conditioning on the current input and insert them into the same experience block used by retrieval-based injection. Unless otherwise noted, the random baseline uses the same number of items or a comparable prompt budget as the corresponding retrieval setting. Global full places a fixed set of full experience items into every request, while Global compact uses a compressed shared summary derived from a much larger experience pool. The schema below summarizes the structure of the compact prompt.

\begin{quote}
\small
\texttt{[Global Compact Experience]}\\
1. When the input contains \{abstract trigger type\}, pay attention to \{abstract risk dimension\}.\\
2. Do not classify \{abstract benign case\} as \{risk label\} unless \{condition\}.\\
3. For ambiguous cases, prefer \{decision principle\}.\\
...
\end{quote}

\paragraph{Reasoning versus Instruct regimes.}
The Instruct and Thinking regimes use the corresponding model settings described in Appendix~\ref{sec:appendix-model-settings}. For Moderation and Tool-use, the final answer is a classification path or a selected tool name, respectively; the Thinking regime additionally allows intermediate reasoning before that final answer. For GPQA, accuracy is evaluated on the final answer option, while completion-token counts cover the generated response in both regimes. The experience block and task-specific instruction follow the same prompt schema across regimes.

\subsection{Global compact and global full variants}
\label{sec:appendix-global-variants}




For the reported global constructions, we count how often each experience item is retrieved from historical online traffic and rank items by this retrieval frequency. This frequency is an exposure-based popularity statistic rather than a human helpfulness score or an evaluation-set performance score. The final evaluation workload is not used to compute these counts.

For Global compact, the 100 experience items with the highest retrieval frequencies are compressed by Claude-opus4.6 into five candidate summaries. Two authors select one candidate using coverage, redundancy, and applicability constraints; the selected prompt is then frozen before evaluation.

\textit{Global full} corresponds to unconditional injection of full experience items. The reported \textit{Global full 50} configuration injects the 50 experience items with the highest historical retrieval frequencies into every request. Thus, Global full and Global compact use the same traffic-derived frequency ranking, but differ in whether the selected experience is injected in full or compressed into a shared summary.

\begin{table*}[t]
\centering
\small
\begin{tabular}{lrrrr}
\toprule
Method & Accuracy (\%) & Full E2E (s) & Selector latency (s) & Select tokens (K) \\
\midrule
Retrieval & 64.7 & 16.92 & 0 & 0 \\
Trigger-aware & 66.7 & 17.58 & 0.76 & 1.04 \\
LLM selector & 78.0 & 76.62 & 58.81 & 55.35 \\
\bottomrule
\end{tabular}
\caption{Measured GPQA Full E2E and model-based selection components. Full E2E includes selection, retrieval, prompt assembly, and generation; selector latency and tokens describe only the selection component. Zero denotes no extra model-based selector, not zero retrieval cost.}
\label{tab:appendix-selector-overhead}
\end{table*}

\begin{table*}[t]
\centering
\small
\begin{tabular}{llcccc}
\toprule
Task & Setting & Accuracy (\%) & Full E2E (s) & Prompt (K) & Completion \\
\midrule
\multirow{2}{*}{Tool-use}
& Retrieval-5 / Instruct & 89.0 & 0.56 & 4.75 & 2.59 \\
& Retrieval-10 / Instruct & 89.1 & 1.29 & 8.86 & 2.58 \\
\midrule
\multirow{2}{*}{Tool-use}
& Retrieval-5 / Thinking & 92.0 & 8.73 & 4.74 & 372.5 \\
& Retrieval-10 / Thinking & 91.6 & 9.51 & 8.89 & 382.8 \\
\bottomrule
\end{tabular}
\caption{Retrieval-depth comparison on tool-use. Moving from Top-5 to Top-10 yields little quality gain while prompt length and latency increase substantially.}
\label{tab:appendix-tooluse-topk}
\end{table*}

\subsection{Selective retrieval, trigger-aware retrieval, and LLM-based selection}
\label{sec:appendix-retrieval-details}

\textit{Retrieval} is the main retrieval-based serving policy studied in this paper. Each experience item is treated as an independent retrieval unit. Given a retrieval query, the system ranks candidate experience items from the task-specific repository and injects only the Top-$K$ selected items into the prompt. Our implementation uses \texttt{bge-m3} for dense semantic retrieval, combines dense and lexical rankings with reciprocal-rank fusion, and reranks a Top-100 candidate pool with \texttt{bge-reranker-base}. The highest-ranked $K$ items form the experience block. Some managed-service stage-level parameters remain unavailable, limiting component-level reproducibility.

Top-$K$ selection uses an independent development set; the final test sets are not used to choose $K$. The main comparisons report Top-$K$=10 for Moderation, Top-$K$=5 for Tool-use, and Top-$K$=10 for standard GPQA Retrieval. Additional test-set sweeps characterize budget sensitivity. Deployment uses Moderation $K=10$, Tool-use $K=5$, and Trigger-aware retrieval for GPQA-like reasoning tasks in agent workflows. These configurations reflect task-specific requirements rather than globally optimal or transferable budgets.

In GPQA, we additionally evaluate \textit{Trigger-aware} retrieval. This variant uses a more targeted query-construction strategy so that retrieval is better aligned with the latent reasoning trigger or knowledge need expressed by the question. The query-construction procedure is described at the strategy level; full system-specific trigger rules are not disclosed. Trigger-aware is also used for GPQA-like reasoning in deployed agent workflows, where latency requirements are less stringent than in Moderation. Its benchmark accuracy difference from standard Retrieval is statistically inconclusive; the deployment choice does not establish a general quality or cost advantage.

We also report an \textit{LLM selector} in GPQA. Here an LLM is used to choose a better experience slice than the default retriever. We treat this setting as a headroom estimate for improved matching rather than as a standard production baseline, since its serving overhead and operational complexity differ substantially from the simpler retriever-based pipeline. Appendix~\ref{sec:appendix-selector-overhead} reports the corresponding selector-overhead comparison.

\begin{table*}[t]
\centering
\scriptsize
\setlength{\tabcolsep}{3pt}
\resizebox{\linewidth}{!}{%
\begin{tabular}{lrrrrrrr}
\toprule
Setting & Prompt (K) & Cache-off E2E (s) & Cache-on E2E (s) & Speedup & Prefix hit (\%) & P95 (s) & QPS \\
\midrule
Global compact & 4.52 & 4.90 & 0.30 & 16.22$\times$ & 99.70 & 0.41 & 65.86 \\
Global full 50 & 18.15 & 18.91 & 0.41 & 46.35$\times$ & 99.90 & 0.55 & 48.11 \\
Retrieval-10 & 3.31 & 1.49 & 0.58 & 2.56$\times$ & 39.90 & 0.68 & 34.12 \\
Retrieval-5 & 2.45 & 1.12 & 0.50 & 2.24$\times$ & 61.20 & 0.54 & 39.52 \\
\bottomrule
\end{tabular}}
\caption{Cache-aware serving comparison on Moderation. Time columns use Full E2E at concurrency 20. Static global prompts benefit from reusable prefixes, while dynamically assembled Retrieval has lower prefix reuse.}
\label{tab:appendix-prefix-cache}
\end{table*}

\begin{table}[t]
\centering
\small
\begin{tabular}{lccc}
\toprule
Setting & Full E2E (ms) & QPS & Completion \\
\midrule
None & 289.01 & 68.76 & 28.26 \\
Global full 50 & 410.64 & 48.11 & 29.78 \\
Global compact & 300.17 & 65.86 & 28.45 \\
Retrieval & 581.26 & 34.12 & 30.52 \\
\bottomrule
\end{tabular}
\caption{Short-output serving-throughput measurements on Moderation. The time column reports Full E2E; QPS measures request throughput. Completion lengths are reported in tokens.}
\label{tab:appendix-shortoutput-throughput}
\end{table}

\section{Qualitative case analysis}
\label{sec:appendix-case-analysis}

This section provides sanitized qualitative illustrations of fine-grained operational guidance and benign over-enforcement, complementing the main-text Moderation diagnostics. These examples illustrate possible behaviors rather than their frequency or a unique causal mechanism.

\subsection{Moderation: retrieval helps on case-specific operational distinctions}

The following cases illustrate how retrieved operational rules can affect policy-conditioned and fine-grained labeling decisions.

\paragraph{Policy-conditioned distinction.}
Consider the query: \textit{``Are drugs necessarily a negative thing?''} In this example, the no-experience model predicts a benign label. Under the operational policy used for annotation, rhetorical questioning that implicitly challenges the legitimacy or negativity of a prohibited category is labeled as policy-relevant risk. Retrieved guidance supplies this policy-specific distinction, illustrating how external experience can support alignment with the operational labeling policy.

\paragraph{Taxonomy-boundary disambiguation.}
Consider the slur-like expression \textit{``Korean stick''}. In this example, the no-experience model assigns a broad insult label. Retrieved guidance distinguishes derogatory expressions tied to protected group identities from generic insults, and the model predicts the discrimination label. This example illustrates how retrieved guidance can support a fine-grained distinction in the operational taxonomy.

These cases illustrate operational distinctions that retrieved guidance can supply beyond the shared taxonomy, including pragmatic rules and adjacent-label boundary cues.

\subsection{Moderation: retrieval can fail through over-triggered or weakly matched rules}

The same mechanism that makes retrieval useful can also create errors when a retrieved rule fires outside its intended applicability boundary. In these cases, the retrieved item is not fully irrelevant; rather, it is a near-match whose trigger conditions are too broad or whose exclusion conditions are missing.

\paragraph{Over-triggered risk pattern.}
Consider the statement: \textit{``I felt exhausted, so I put my head down on the desk and cried, but no one noticed.''} The no-experience model correctly treats this as an ordinary expression of sadness or fatigue. Retrieval, however, can return mental-health-risk precedents whose surface cues overlap with the input, such as loneliness or emotional invisibility. When those cues are applied too aggressively, the model over-classifies ordinary distress as a higher-risk mental-health case.

\paragraph{Over-generalized structural cue.}
Consider the question: \textit{``If someone comes from a poor family background, will that affect dating or relationships?''} The no-experience model again predicts a benign label. Retrieval can misfire by activating an overly broad structural rule that maps questions about social background and outcomes into discrimination-related labels. Here the error comes from applying a rule intended for group-level derogatory attribution to a much broader class of ordinary social questions.

These cases illustrate how near-matching rules can cause over-triggering and boundary errors.

\section{Selector-overhead comparison on GPQA}
\label{sec:appendix-selector-overhead}

Table~\ref{tab:appendix-selector-overhead} reports measured Full E2E together with the model-based selection components for GPQA. Trigger-aware has Full E2E of 17.58\,s, including 0.76\,s of selection, while the LLM selector has Full E2E of 76.62\,s, including 58.81\,s of selection. These complete-path values are measured directly rather than inferred from selector timings. The Trigger-aware accuracy difference is not statistically conclusive; selector-stage latency and tokens are component measurements, not substitutes for Full E2E.

\section{Retrieval-depth comparison on tool-use}
\label{sec:appendix-tooluse-topk}

Table~\ref{tab:appendix-tooluse-topk} reports the retrieval-depth comparison for tool-use, complementing the moderation Top-$K$ analysis in the main text. The pattern is consistent with the Moderation results: increasing retrieval depth beyond a moderate value yields little additional quality gain, while Full E2E and prompt length continue to increase substantially.

\section{Cache-aware serving comparison}
\label{sec:appendix-prefix-cache}

Table~\ref{tab:appendix-prefix-cache} reports the cache-aware serving comparison referenced in the main text. The results show that static global prompts benefit more from reusable prefix caching than dynamically assembled retrieval prompts, although retrieval remains competitive because its prompt payload is smaller.

\section{Short-output serving-throughput measurements}
\label{sec:appendix-shortoutput-throughput}

Table~\ref{tab:appendix-shortoutput-throughput} reports additional short-output serving measurements that complement the token and Full E2E comparisons in the main paper. These results provide a more deployment-oriented view of the same trade-off through Full E2E and QPS on a Moderation short-output serving baseline.

\section{Additional workload and budget measurements}
\label{sec:appendix-workload-budget}
Table~\ref{tab:appendix-interface-details} reports the full Instruct serving-interface measurements, including prompt and completion lengths. Table~\ref{tab:appendix-regime-matrix} reports detailed Instruct/Thinking measurements. Table~\ref{tab:appendix-gpqa-8b-control} reports the same-generator GPQA control. These are configuration-specific comparisons, not a full factorial isolation of output structure.

\begin{table*}[t]
\centering
\small
\setlength{\tabcolsep}{4pt}
\begin{tabular}{llrrrr}
\toprule
Task & Setting & Acc. (\%) & Full E2E (s) & Prompt (K) & Compl. \\
\midrule
\multirow{5}{*}{Moderation}
& None & 46.7 $\pm$ 0.2 & 0.29 & 1.26 & 28.26 \\
& Random & 57.0 $\pm$ 0.4 & 0.29 & 4.12 & 27.57 \\
& Global compact & 69.4 $\pm$ 0.3 & 0.30 & 4.52 & 28.45 \\
& Global full 50 & 63.1 $\pm$ 0.2 & 0.41 & 18.1 & 29.78 \\
& Retrieval & 82.1 $\pm$ 0.1 & 0.58 & 3.31 & 30.52 \\
\midrule
\multirow{5}{*}{Tool-use}
& None & 67.9 $\pm$ 0.1 & 0.08 & 0.52 & 2.79 \\
& Random & 69.5 $\pm$ 0.1 & 0.09 & 2.89 & 2.74 \\
& Global compact & 71.9 $\pm$ 0.2 & 0.12 & 1.92 & 2.70 \\
& Global full 50 & 72.9 $\pm$ 0.2 & 0.25 & 23.48 & 2.69 \\
& Retrieval & 89.0 $\pm$ 0.1 & 0.56 & 4.75 & 2.59 \\
\midrule
\multirow{5}{*}{GPQA-diamond}
& None & 61.9 $\pm$ 0.2 & 19.35 & 0.27 & 2.44K \\
& Random & 62.0 $\pm$ 0.3 & 23.16 & 2.04 & 2.61K \\
& Global compact & 62.8 $\pm$ 0.5 & 25.08 & 2.91 & 2.62K \\
& Global full 50 & 61.5 $\pm$ 0.2 & 33.26 & 11.08 & 2.65K \\
& Retrieval & 64.7 $\pm$ 0.4 & 16.92 & 1.85 & 1.42K \\
\bottomrule
\end{tabular}
\caption{Detailed Instruct serving-interface measurements underlying Table~\ref{tab:main-comparison}. Accuracy is reported as the mean $\pm$ standard deviation over three runs for Moderation and Tool-use and over eight runs for GPQA-diamond. Prompt lengths are in thousands of tokens; completion lengths are in tokens, with K denoting thousands.}
\label{tab:appendix-interface-details}
\end{table*}

\begin{table*}[t]
\centering
\small
\begin{tabular}{lllcccccc}
\toprule
Task & Regime & Inference & Setting & Acc. (\%) & Full E2E (s) & Prompt (K) & Compl. & $\Delta$Acc. (pp) \\
\midrule
\multirow{4}{*}{Moderation}
& \multirow{4}{*}{Short-output} & Instruct & None & 46.7 $\pm$ 0.2 & 0.29 & 1.26 & 28.3 & -- \\
&  & Instruct & Retrieval & 82.1 $\pm$ 0.1 & 0.58 & 3.31 & 30.5 & +35.4 \\
&  & Thinking & None & 58.1 $\pm$ 0.2 & 3.08 & 1.26 & 401.8 & -- \\
&  & Thinking & Retrieval & 85.6 $\pm$ 0.3 & 3.92 & 3.31 & 475.0 & +27.5 \\
\midrule
\multirow{4}{*}{Tool-use}
& \multirow{4}{*}{Short-output} & Instruct & None & 67.9 $\pm$ 0.1 & 0.08 & 0.52 & 2.8 & -- \\
&  & Instruct & Retrieval & 89.0 $\pm$ 0.1 & 0.56 & 4.75 & 2.6 & +21.1 \\
&  & Thinking & None & 77.9 $\pm$ 0.3 & 4.40 & 0.52 & 288.1 & -- \\
&  & Thinking & Retrieval & 92.0 $\pm$ 0.2 & 8.73 & 4.74 & 372.5 & +14.1 \\
\midrule
\multirow{4}{*}{GPQA}
& \multirow{4}{*}{Decode-heavy} & Instruct & None & 61.9 $\pm$ 0.2 & 19.35 & 0.27 & 2.44K & -- \\
&  & Instruct & Retrieval & 64.7 $\pm$ 0.4 & 16.92 & 1.85 & 1.42K & +2.8 \\
&  & Thinking & None & 68.3 $\pm$ 0.2 & 72.55 & 0.25 & 6.31K & -- \\
&  & Thinking & Retrieval & 70.0 $\pm$ 0.2 & 63.09 & 1.91 & 5.12K & +1.7 \\
\bottomrule
\end{tabular}
\caption{Detailed cross-task measurements for the Instruct and Thinking comparisons. Accuracy is reported as the mean $\pm$ standard deviation over three runs for Moderation and Tool-use and over eight runs for GPQA-diamond; $\Delta$Acc. denotes Retrieval minus None within the same inference regime.}
\label{tab:appendix-regime-matrix}
\end{table*}

\begin{table*}[t]
\centering
\small
\begin{tabular}{lrrr}
\toprule
Qwen3-8B Thinking / GPQA & Accuracy (\%) & Completion tokens & Full E2E (s) \\
\midrule
None & $61.62\pm0.41$ & 7795.5 & 106.017 \\
Retrieval & $66.86\pm0.36$ & 6384.3 & 96.341 \\
\bottomrule
\end{tabular}
\caption{Qwen3-8B Thinking GPQA control; Accuracy is reported as the mean $\pm$ standard deviation over eight runs. Retrieval shortens completion and lowers Full E2E in this configuration. Moderation and Tool-use also use Qwen3-8B, although task prompts and repositories differ.}
\label{tab:appendix-gpqa-8b-control}
\end{table*}

\begin{table*}[t]
\centering
\small
\begin{tabular}{llrrr}
\toprule
Task / Instruct & Budget & Accuracy (\%) & Full E2E (s) & Prompt (K) \\
\midrule
Moderation & K0 (None) & 46.7 & 0.29 & 1.26 \\
Moderation & K1 & 65.8 & 0.44 & -- \\
Moderation & K3 & 72.1 & 0.48 & -- \\
Moderation & K5 & 76.4 & 0.50 & -- \\
Moderation & K10 & 82.1 & 0.58 & 3.31 \\
Moderation & K20 & 81.3 & 0.70 & -- \\
Moderation & K30 & 79.2 & 0.81 & 6.66 \\
Tool-use & K5 & 89.0 & 0.56 & 4.75 \\
Tool-use & K10 & 89.1 & 1.29 & 8.86 \\
GPQA & K10 & 64.7 & 16.92 & 1.85 \\
\bottomrule
\end{tabular}
\caption{Tabulated retrieval-budget measurements used in the main-text analysis. Figure~\ref{fig:quality-cost} additionally shows Tool-use $K=1$ and $K=3$ and GPQA $K=5$ and $K=20$ operating points. A dash denotes an unreported matched measurement. Moderation sweep latencies are reported separately from the cache benchmark in Table~\ref{tab:appendix-prefix-cache}.}
\label{tab:appendix-budget-points}
\end{table*}

\end{document}